\documentclass{article} 
\usepackage[dvipsnames]{xcolor}
\usepackage{iclr2025_conference,times}

\usepackage{amsmath,amsfonts,bm}

\def\eqref#1{equation~\ref{#1}}

\def\1{\bm{1}}

\DeclareMathAlphabet{\mathsfit}{\encodingdefault}{\sfdefault}{m}{sl}
\SetMathAlphabet{\mathsfit}{bold}{\encodingdefault}{\sfdefault}{bx}{n}

\newcommand{\magmax}{\textsc{M\MakeLowercase{ag}M\MakeLowercase{ax}}}

\usepackage[utf8]{inputenc} 
\usepackage[T1]{fontenc}    

\usepackage{multirow}
\usepackage{graphicx}
\usepackage{epstopdf}
\usepackage{booktabs}
\usepackage{array}
\usepackage{wrapfig}
\usepackage{adjustbox}
\usepackage{subfigure}
\usepackage{graphicx,subcaption}
\usepackage{hyperref}
\usepackage{cleveref}
\usepackage{url}

\title{Low-Rank Continual Personalization \\ of Diffusion Models}

\vspace{-2em}
\author{Łukasz Staniszewski\thanks{Equal Contribution.}\\
  Warsaw University of Technology\\
  \texttt{luks.staniszewski@gmail.com} \\
  \vspace{-2em}
  \And
  Katarzyna Zaleska$^*$\\
  Warsaw University of Technology \\
  \texttt{katarzyna.zaleska2.stud@pw.edu.pl} \\
    \vspace{-2em}
  \AND
  Kamil Deja \\
  Warsaw University of Technology \\
  IDEAS NCBR \\
  \texttt{kamil.deja@pw.edu.pl} \\
}
\vspace{-8em}

\iclrfinalcopy 
\begin{document}

\maketitle

\vspace{-1em}
\begin{abstract}
\vspace{-0.5em}
Recent personalization methods for diffusion models, such as Dreambooth and LoRA, allow fine-tuning pre-trained models to generate new concepts. However, applying these techniques across consecutive tasks in order to include, e.g., new objects or styles, leads to a forgetting of previous knowledge due to mutual interference between their adapters. In this work, we tackle the problem of continual customization under a rigorous regime with no access to past tasks' adapters. In such a scenario, we investigate how different adapters' initialization and merging methods can improve the quality of the final model. To that end, we evaluate the na\"ive continual fine-tuning of customized models and compare this approach with three methods for consecutive adapters' training: sequentially merging new adapters, merging orthogonally initialized adapters, and updating only relevant task-specific weights. In our experiments, we show that the proposed techniques mitigate forgetting when compared to the na\"ive approach. In our studies, we show different traits of selected techniques and their effect on the plasticity and stability of the continually adapted model. 
Repository with the code is available at \href{https://github.com/luk-st/continual-lora}{\color{RubineRed}{https://github.com/luk-st/continual-lora}}.

\end{abstract}

\section{Introduction}

Diffusion models~\citep{sohl2015deep} (DMs) have revolutionized image generation with state-of-the-art performance in synthesizing detailed objects and blending them with complex styles. Personalization techniques such as Dreambooth~\citep{dreambooth}, combined with the Low-Rank Adaptation (LoRA, \cite{lora}), introduce the possibility to easily customize large-scale pre-trained models. In those techniques, a selected subset of model parameters is updated with just a few training examples to enable the generation of concepts not present in the original training set.

Unfortunately, customization of a generative model over several tasks leads to the catastrophic forgetting~\citep{1999french} of previous knowledge due to mutual interference between tasks' adapters. Recent studies~\citep{ziplora, blora} try to tackle this issue in terms of object and style customization tasks' adapters by carefully merging the trained adapters, either with optimization~\citep{ziplora} or by targeting different attention layers~\citep{blora}. However, for the merging to succeed, the adapters have to be available at the same time to properly scale~\citep{ziplora} or select~\citep{marczak2024magmax} the resulting models' weights. Even though LoRA reduces the number of adapted parameters, in practice, with many objects or styles, it is unfeasible to store copies of model weights for each of them. Even in the form of sparse matrices, keeping all the adapters is memory-consuming (e.g., approximately 8 LoRA weight matrices with rank $64$ are equal to the size of all adapted parameters in Stable Diffusion XL's~\citep{stablediffusionxl2024} model). 

Therefore, in this work, we propose to study the effectiveness of LoRA weights initialization and merging \textit{under the strict continual learning regime} where only the model or model with a single adapter is passed between tasks. In a series of experiments, we compare (1) Na\"ive continual fine-tuning of the low-rank adapter, and three simple merging baselines mitigating forgetting of concepts and styles: (2) consecutive merging of task-specific adapters, (3) merging LoRAs initialized with orthogonal weights, and (4) merging through a selection of weights with the highest magnitude for the task. Our experiments indicate that adding multiple adapters in a Na\"ive way can lead to a situation where a model converges, in its performance, towards its base form, while all the evaluated techniques mitigate this issue. 

We show the significant distinctions between different initialization methods. For example, orthogonalized initialization results in high plasticity but low stability, whereas selection-based merging exhibits the opposite trend. Our findings indicate that optimal performance is achieved by reinitializing the LoRA weights and merging them using a standard approach. We hypothesize that this effect arises from the specific initialization of LoRA weights, which facilitates sparse adaptation of the base model weights.
\section{Related work}
\paragraph{Diffusion models personalization and customization}
The topic of Diffusion Models personalization has been broadly studied in recent years. Several works, including Textual Inversion~\citep{gal2022image} learn new word embeddings to represent custom and unique concepts. This idea is further extended by~\citet{li2023blipdiffusionpretrainedsubjectrepresentation,kumari2022multiconcept}, with additional regularization~\citep{wu2024core}, or features mapping~\citep{wei2023elite}. On the other hand, DreamBooth~\citep{dreambooth} fine-tunes the entire model to incorporate specific subjects into generated images. This idea is further extended in HyperDreamBooth~\citep{ruiz2024hyperdreambooth} where a hypernetwork~\citep{ha2016hypernetworks} is used to generate personalized weights. 
Recent personalization techniques employ Low-Rank Adaptation to reduce the computational requirements by fine-tuning small, low-rank matrices and keeping the base model frozen. Nevertheless, the simple merging of weights fine-tuned with LoRA does not prevent task interference. Therefore, in ZipLoRA~\citet{ziplora} optimize separately trained adapters for style and object to simplify their merging, and \citet{blora} (B-LoRA) target different attention layers of diffusion models for both style and object adapters. OFT~\citep{Qiu2023OFT} and Orthogonal Adaptation~\citep{Po_2024_CVPR} leverage orthogonalization to merge models seamlessly without compromising fidelity. Finally, \citet{dravid2024interpreting} introduce weights2weights -- a personalized adapter space, enabling the creation of entirely new personalized models. 

\paragraph{Generating style with diffusion models}
Apart from introducing new objects to the model, there is a line of research aiming for style generation and editing in Diffusion Models. StyleDrop~\citep{sohn2023styledrop} leverages the Muse transformer to fine-tune text-to-image models for generating style-specific images. StyleAligned~\citep{hertz2023StyleAligned} employs shared attention to maintain a consistent style across generated images, while Style Similarity~\citep{somepalli2024measuring} provides a framework for style retrieval from diffusion models' training data and introduces CSD, an image encoder that accurately captures the style of images. UnlearnCanvas~\citep{unlearn-dataset} offers a comprehensive dataset for evaluating machine unlearning in stylized image generation.
Diff-QuickFix~\citep{localizingandediting} and LOCOEDIT~\citep{onmechanisticknowledge} enable precise editing of diffusion models' knowledge of specific styles by identifying style-relevant layers.

\paragraph{Continual Learning of Diffusion Models}
There is a growing line of research employing DMs in continual learning (CL) mainly as a source of rehearsal examples from past tasks~\citep{gao2023ddgr}. This idea is extended by~\cite{cywinski2024guide} through classifier guidance towards easy-to-forget examples. Similarly, \cite{jodelet2023classincremental} uses an externally trained text-to-image diffusion model for the same purpose.
On the other hand, \cite{zajac2023exploring} explore if contemporary CL methods prevent forgetting in diffusion models.

\citet{smith2023continual} introduce the task of continual customization, with the goal of fine-tuning the diffusion model on consecutive personalization tasks. They introduce C-LoRA, where consecutive tasks are learned with a self-regularization technique that penalizes updates corresponding to the already altered weights. STAMINA~\citep{smith2024continual} introduces learnable adapter's attention masks parameterized with low-rank MLPs for better scaling to longer task sequences. CIDM~\citep{dong2024how} develop a concept-consolidation loss that explores both the distinctive characteristics of concepts within each task (Task-Specific Knowledge) and shared information across different tasks with similar concepts (Task-Shared Knowledge). Finally, \citet{jha2024mining} propose to leverage classifier scores into both parameter-space (to improve the task-specific Fisher information estimates of EWC) and function-space regularizations (creating a double-distillation framework), achieving high performance in continual personalization. While prior methods improve knowledge consolidation using additional regularizations, in this work, we study the fundamental properties of the most basic components for continual personalization which are merging and initialization techniques.

\paragraph{Model merging}
The concept of editing models through arithmetic operations applied on models' weights is explored by \citet{task_arithmetic}, where authors describe mathematical operations between task vectors, defined as the difference between the weights of the fine-tuned and the base model. To limit the interference between tasks, recent methods replace straightforward averaging with summation of trimmed parameters~\citep{yadav2024ties}, Fisher-information-based merging~\citep{matena2022merging} or with guidance from discriminative model~\citep{jin2022dataless}. In this work, we adapt $\magmax$~\citep{marczak2024magmax} to our experimental setting, the method achieving state-of-the-art performance in both class-incremental and domain-incremental learning in discriminative modeling. $\magmax$ sequentially fine-tunes the model and consolidates the knowledge thanks to maximum magnitude weight selection from several low-rank task adapters.
\section{Methods}\label{sec:methods}

\begin{figure}
    \centering
    \begin{minipage}{0.44\linewidth}
        \includegraphics[width=\linewidth]{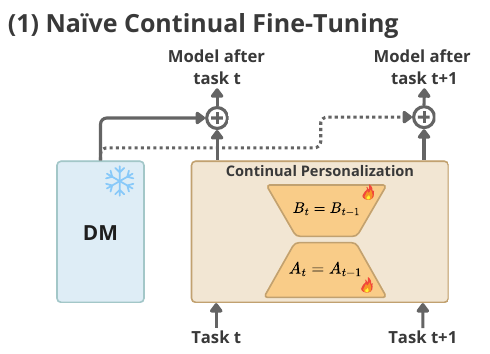}
    \end{minipage}
    \hspace{0.05\linewidth}
    \begin{minipage}{0.44\linewidth}
        \vspace{-0.6em}
        \includegraphics[width=\linewidth]{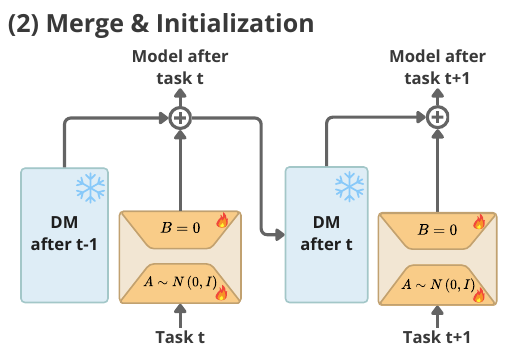}
    \end{minipage}
    \begin{minipage}{0.44\linewidth}
        \includegraphics[width=\linewidth]{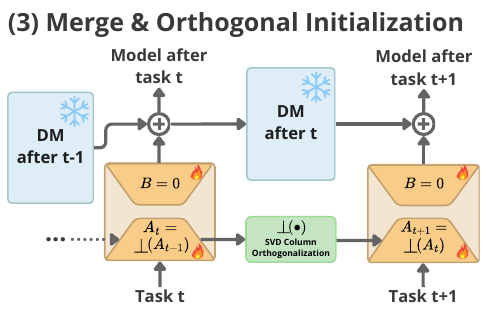}
    \end{minipage}
    \hspace{0.05\linewidth}
    \begin{minipage}{0.44\linewidth}
        \vspace{-0.6em}
        \includegraphics[width=\linewidth]{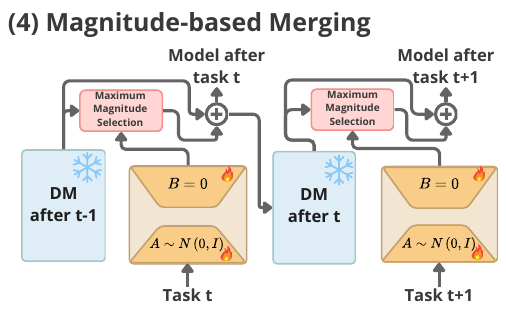}
    \end{minipage}
\caption{Comparison of initialization and merging in four evaluated methods. The expressions in the adapters refer to how they are initialized with a new task. Numbers next to each schema refer to textual descriptions in~\Cref{sec:methods}. \vspace{-0.5cm}}
\label{fig:experiments_setup}
\end{figure}

We evaluate four different approaches to continual diffusion model personalization. Notation-wise, we denote $W_0$ as the base model weights, while $\{A_{t}, B_{t}\}$ represent the fine-tuned low-rank adapter resulting from personalizing DM on the $t$-th task. New adapter weights $\{A_{t}, B_{t}\}$ are calculated through fine-tuning on data from task $t$, given the frozen base model weights $W_0$, and an adapter fine-tuned over $t-1$ tasks denoted as $\{A_{1\dots t-1}, B_{1\dots t-1}\}$. The final model can be calculated as $W_{1\dots T} = W_{0} + B_{1\dots T}A_{1\dots T}$.
All the evaluated strategies differ in either initialization or merging strategy, which we overview in Fig.~\ref{fig:experiments_setup}, and describe further in this section.

\paragraph{(1) Na\"ive continual fine-tuning of LoRA adapters.}
In the na\"ive approach, we sequentially fine-tune the adapter weights from the previous task $\{A_{t-1},B_{t-1}\}$ with the current task data without any reinitialization. Thus, the intermediate adapters, which are continually changed, can be merged into the base model after completing all the $T$ tasks.

\paragraph{(2) Continual adapters merging and standard LoRA reinitialization.}
We create a new LoRA adapter for each task with a standard initialization technique ($A \sim \mathcal{N}(0,\mathcal{I}), B = 0$). Then, at the end of each task, we merge the adapter with the base model and pass it to the next task. This approach is similar to na\"ive finetuning. However, thanks to the reinitialization of the LoRA parameters with random values for the $A$ matrix, each task is optimized in its own independent low-dimensional space.
Therefore, compared to a naïve approach, the Merge \& Initialization technique naturally reduces task interference, as each adapter independently learns only a minimal set of task-specific parameters. \cite{marczak2025task} show that such situation has positive impact on the final model perfromance in discriminative modeling.
Intuitively, this can be related to the two-stage training~\citep{kamra2017deep} introduced for various CL methods.

\vspace{-0.5em}
\paragraph{(3) Continual adapters merging and orthogonalized reinitialization.} 
We further extend the merging and reinitialization approach in order to limit the interference between tasks by initializing, for the given task $t$, $B_t$ weights with zeros and setting $A_t$ weights' columns as orthogonal to columns of $A_{1\dots t-1}$ matrix using Singular Value Decomposition (SVD). We decompose the $i$-th column of $A_{1\dots t-1}=\Sigma_{s\in1\dots t-1}A_s$ as
\vspace{-0.5em}
\begin{equation}
    {A^{(i)}_{1\dots t-1}} = \mathbf{U} \mathbf{\Sigma} \mathbf{V}^H
\vspace{-0.3em}
\end{equation}
and select the last row of the conjugate transpose of the right singular vectors $\mathbf{V}$, corresponding to the smallest singular value. The underlying assumption of this approach is to pre-direct the fine-tuning process toward the non-conflicting adapters, thus limiting the mutual drift of weights that are introduced by consecutive tasks.

\vspace{-0.5em}
\paragraph{(4) Magnitude-based selection of LoRA weights.}
Finally, we adapt and \textit{modify} the magnitude-based parameter selection method introduced in~\cite{marczak2024magmax}. As the original $\magmax$ approach involves merging adapters into the model after all the $T$ tasks, continuing the method for $(T+1)$-th task is impossible without storing base model weights $\theta_0$, weights obtained after $T$ tasks $\theta_T$ (before selection operation), and the running max-magnitude statistics $\tau_\magmax$. To that end, after \textit{each task} $t$, we run the $\magmax$ selection of parameters with the highest magnitude by comparing the sum of already merged adapters $\{A_{1\dots t-1}, B_{1\dots t-1}\}$ with the current one $\{A_{t}, B_{t}\}$:
\vspace{-0.1em}
\begin{equation}\label{eq:magmax}
    \{A_{1\dots t},B_{1\dots t}\} = \{ \magmax(A_{1\dots t-1};A_t),\magmax(B_{1\dots t-1};B_t)\}.
\vspace{-0.1em}
\end{equation}
The resulting adapter is passed together with a base model into the next task. At the beginning of a new task, the so far selected adapters are temporarily merged into the frozen Base model, while the new adapters' parameters are initialized in the standard way, optimized, and passed to the selection procedure (\ref{eq:magmax}).

\section{Experiments} 

\paragraph{Experimental setup.} In our experiments, we use the DreamBooth~\citep{dreambooth} dataset for fine-tuning on objects, and  UnlearnCanvas~\citep{unlearn-dataset} for styles. 
We perform fine-tuning on the $10$ consecutive tasks and average results over 4 random tasks ordering, with 2 random seeds each. To measure the performance of the model, we calculate cosine similarity between model outputs and reference images with DINO~\citep{dino} and CLIP~\citep{clip} feature extractors for object alignment measure and DINO and CSD~\citep{csd} for style similarity. We report the performance of the final model fine-tuned over all 10 tasks with the Average Score and Average Forgetting metrics. More details about the experimental setup are provided in the~\Cref{appendix:experimental-setup}. 

\vspace{-0.5em}
\paragraph{Results.}
In \Cref{table:object_comparison} and \Cref{table:style_comparison}, we present the results for the continual object and style personalization, respectively. For both tasks, we observe that the merge \& initialization method, either with or without orthogonalization, outperforms na\"ive continual fine-tuning. On the other hand, magnitude-based merging has the lowest average forgetting at the expense of lower general performance as a result of the lower adaptation capabilities of the method. While magnitude-based weights selection yields state-of-the-art performance in discriminative tasks~\citep{marczak2024magmax}, selecting a limited number of weights hinders the precise merging of new generative traits. Detailed analysis in~\Cref{appendix:results} on performance after each task shows that the merge \& orthogonalization approach exhibits the best plasticity. However, a simple summation of orthogonal task vectors makes Merge \& Orthogonalization fail in retaining original performance on past tasks, leading to the highest forgetting. The Merge \& Initialization method balances well a relatively high plasticity with stability, as observed by well-preserved past knowledge.

\begin{table*}[!ht]
\centering
\small
\begin{tabular}{@{}lcccccc@{}}
\toprule
& \multicolumn{2}{c}{\textbf{Average Score} $\overline{S}_T \, (\uparrow)$}  & \multicolumn{2}{c}{\textbf{Average Forgetting} $\overline{F}_T \, (\downarrow)$} \\
\cmidrule{2-5}

\textbf{Adapter fine-tuning method} & \textbf{CLIP-I} & \textbf{DINO} & \textbf{CLIP-I} & \textbf{DINO} \\ 
\midrule
Base model (reference) & $0.586$  & $0.304$ & - & - \\
\midrule
Na\"ive Continual Fine-Tuning  & $0.670_{\pm .013}$  & $0.402_{\pm .029}$ & $0.063_{\pm .013}$ & $0.144_{\pm .039}$ \\
Merge \& Initialization  & $\boldsymbol{0.675}_{\pm .005}$  & $\boldsymbol{0.457}_{\pm .011}$ & $\underline{0.026}_{\pm .003}$  & $\underline{0.056}_{\pm .009}$ \\
Merge \& Orthogonal Initialization  & $\underline{0.673}_{\pm .014}$ & ${0.403}_{\pm .025}$ & $0.072_{\pm .015}$ & $0.162_{\pm .033}$ \\
Magnitude-based Merging  & $0.643_{\pm .002}$ & $\underline{0.408}_{\pm .006}$ & $\boldsymbol{0.018}_{\pm .002}$ & $\boldsymbol{0.036}_{\pm .006}$ \\
\bottomrule
\end{tabular}
\caption{Average Score and Average Forgetting in CLIP-I and DINO alignment metrics for continual object personalization. The best results are in \textbf{bold}, while the second best are \underline{underlined}.}
\label{table:object_comparison}
\end{table*}

\begin{table*}[!ht]
\small
\centering
\begin{tabular}{@{}lcccccc@{}}
\toprule
& \multicolumn{2}{c}{\textbf{Average Score} $\overline{S}_T \, (\uparrow)$}  & \multicolumn{2}{c}{\textbf{Average Forgetting} $\overline{F}_T \, (\downarrow)$} \\
\cmidrule{2-5}
\textbf{Adapter fine-tuning method} & \textbf{CSD} & \textbf{DINO} & \textbf{CSD} & \textbf{DINO} \\ 
\midrule
Base model (reference) & $0.088$  & $0.146$ & - & - \\
\midrule
Na\"ive Continual Fine-Tuning  & $0.345_{\pm .032}$ & $0.249_{\pm .010}$ & $0.184_{\pm .039}$ & $0.136_{\pm .013}$ \\
Merge \& Initialization  & $\boldsymbol{0.385}_{\pm .019}$ & $\boldsymbol{0.285}_{\pm .019}$ & $\underline{0.131}_{\pm .019}$ & $\underline{0.085}_{\pm .021}$ \\
Merge \& Orthogonal Initialization  & $\underline{0.349}_{\pm .014}$ & $\underline{0.252}_{\pm .012}$ & $0.204_{\pm .013}$ & $0.141_{\pm .013}$ \\
Magnitude-based Merging  & $0.289_{\pm .015}$ & $0.240_{\pm .009}$ & $\boldsymbol{0.093}_{\pm .013}$ & $\boldsymbol{0.052}_{\pm .015}$ \\
\bottomrule
\end{tabular}
\caption{Average Score and Average Forgetting in CSD and DINO alignment metrics for continual style personalization. The best results are in \textbf{bold}, while the second best are \underline{underlined}. \vspace{-0.4cm}}
\label{table:style_comparison}
\end{table*}

\begin{figure}[h!]
    \centering
    \subfigure[Object personalization]{\includegraphics[width=0.4\textwidth]{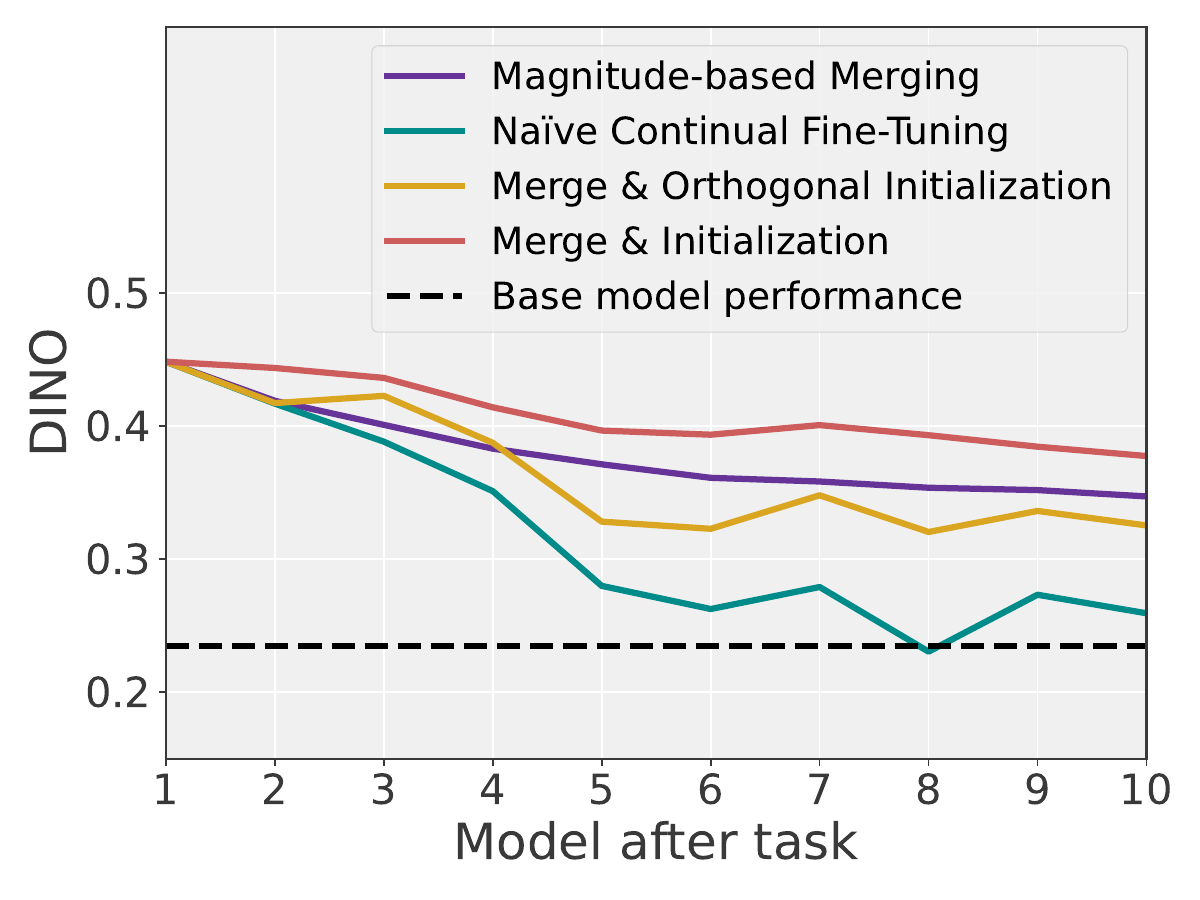}}
    \hspace{0.05\textwidth}
    \subfigure[Style personalization]{\includegraphics[width=0.4\textwidth]{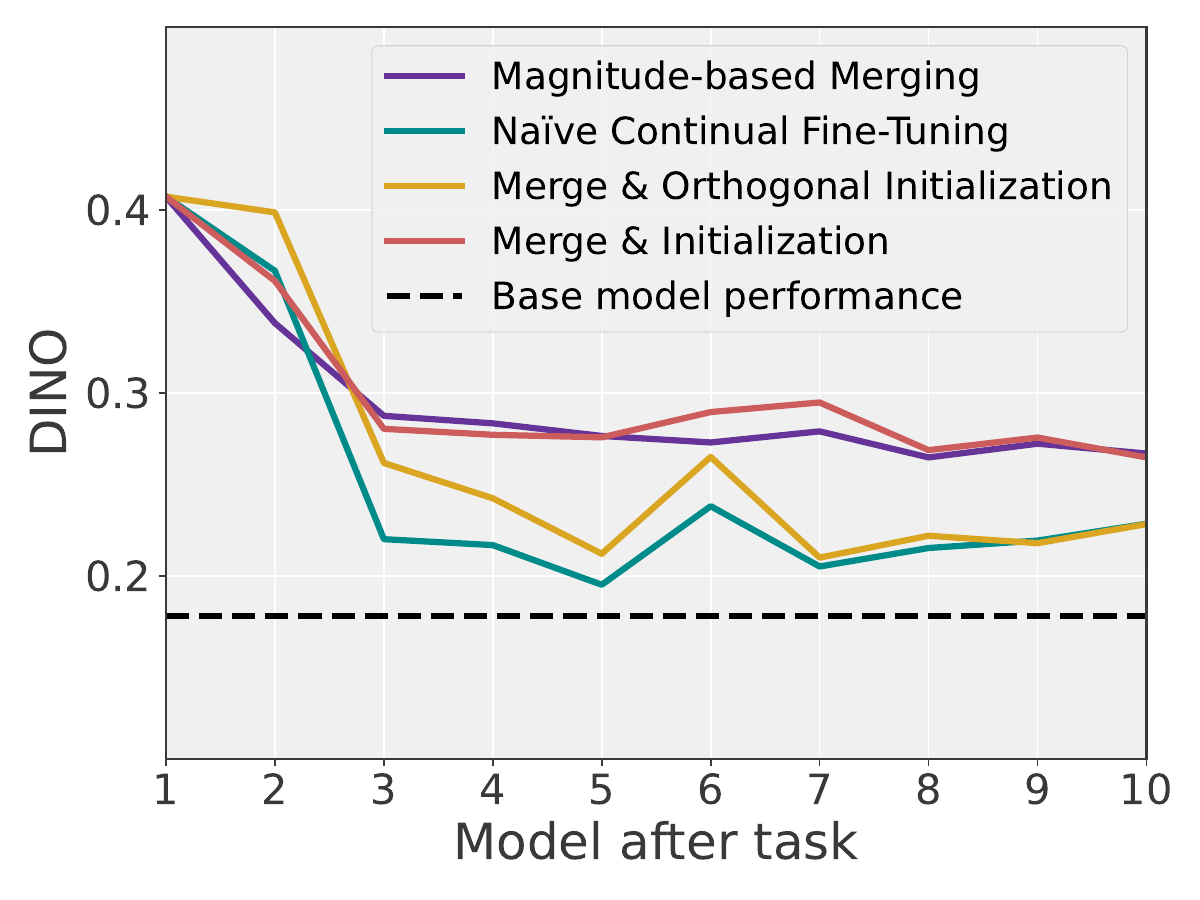}}
\caption{\vspace{-0.1cm}DINO score on the first task over continual fine-tuning on next tasks for the object and styles customization.}
\label{fig:task1_and_ratio_objects}
\end{figure}

In Fig.~\ref{fig:task1_and_ratio_objects}, we compare the performance of different methods on the first task after continual adaptation across multiple tasks in object (left) and style (right) personalization. The results show that the na\"ive approach degrades overall quality to the baseline model’s performance, while other techniques mitigate forgetting more effectively. Notably, knowledge retention varies significantly between object and style personalization. While the Merge \& Initialization method reduces forgetting in both cases, resetting weights with orthogonal initialization performs significantly worse in style personalization. We hypothesize that this is due to task vectors being more aligned in style personalization. In such cases, more orthogonal vectors lead to higher forgetting.

\section{Conclusions}

In this work, we analyze the continual adaptation capabilities of a DreamBooth method coupled with the Low-Rank Adaptation (LoRA) technique. We evaluate three different approaches to merging and initialization of the individual adapters. We show that na\"ive continual training of a LoRA leads to catastrophic forgetting, while other techniques can mitigate this issue.

\section*{Acknowledgments}
The project was funded by the National Science Centre, Poland, grants no: 2023/51/B/ST6/03004 and 2022/45/B/ST6/02817. The computational resources were provided by PLGrid grant no. PLG/2024/017114 and no. PLG/2024/017266.

\bibliography{iclr2025_conference}
\bibliographystyle{iclr2025_conference}

\newpage
\appendix
    \section*{\LARGE Appendix}

\section{Experimental setup}
\label{appendix:experimental-setup}
\subsection{Datasets}

\textbf{DreamBooth.} To fine-tune the model on personalized objects, the DreamBooth~\citep{dreambooth} dataset is used, which contains images of 30 objects (9 of them being dogs and cats, and 21 being other objects) belonging to 15 different classes. Each object is represented by several images (from 4 up to 6), which represent it in either a different environment or from a different camera angle. Due to the quantity limits, we follow \cite{dreambooth} and use the same dataset of images for both evaluation and fine-tuning.

\textbf{UnlearnCanvas.} For style-related tasks, we use the UnlearnCanvas collection~\citep{unlearn-dataset}, which contains 20 objects (each with 20 images) in 60 different styles. When evaluating methods for the style tasks, we select $10$ images representing $10$ different objects in the same style but the ones that were not used during training.

We generate datasets for our experiments by randomly selecting classes and objects. For object-related tasks, we select 10 classes, each represented by a unique object and paired with a unique token. For style-based tasks, we select 10 distinct styles, each paired with one object and five corresponding images. Object names, classes, and styles were uniquely selected to avoid repetition. Selected classes are presented in~\Cref{table:style_dataset}.

\begin{table*}[!h]
\centering
\begin{tabular}{@{}p{3cm}p{10cm}@{}}
\toprule
\textbf{Style name} & byzantine, palette knife, ink art, blossom season, artist sketch, \newline winter, dadaism, dapple, meteor shower, pastel \\ 
\midrule
\textbf{Object name} & sneaker, boot, backpack, stuffed animal, candle, can, glasses, bowl, cartoon, teapot \\ 
\bottomrule
\end{tabular}
\caption{Objects and styles names used during experiments.}
\label{table:style_dataset}
\end{table*}

\subsection{Finetuning}

We perform the experiments in a class-incremental setup, as illustrated in~\Cref{fig:experiments_setup}. Starting with a base text-to-image Stable Diffusion XL model, we sequentially train it on the following tasks. After each task, depending on the approach, we either merge the obtained LoRA weights with the base model and pass it to the next task or store only the LoRA weights to initialize adapters for the subsequent steps.

We train the model on $T=10$ tasks. For each object task, we use distinct DreamBooth tokens, while for styles, we retain the real names. The learning rate is set to 1e-5, with 850 train steps and a batch size of 1. During fine-tuning, we don't leverage the prior preservation regularization~\cite{dreambooth}.

In the original work introducing LoRA~\citep{lora}, the authors investigate how both the rank and the target matrices of the transformer's self-attention module, to which the LoRA is applied, influence method efficiency. They show that, while the impact of rank $r$ is small, it is recommended to apply the LoRA to $W^Q$, $W^K$, $W^V$, and $W^O$. In our experiments, we similarly apply LoRA to these targets within both the self-attention and cross-attention layers of the U-Net and set $r=64$.

To ensure the robustness of our experiments, all four methods for low-rank continual personalization methods were evaluated with two seeds ($0$, $5$) and four task ordering seeds ($0$, $5$, $10$, $42$).

\newpage
\subsection{Evaluation metrics}

To evaluate fine-tuned model generations, we use image alignment metrics to calculate the cosine similarity between embeddings of generated and reference images. For subject alignment, we use features extracted from the DINO~\cite{dino} and CLIP~\cite{clip} models. When assessing style, we continue using DINO but replace CLIP with CSD~\cite{csd}, as CLIP has demonstrated a bias towards image content, making it unsuitable for style-related tasks. When sampling from the text-to-image diffusion model, we generate $N=8$ images from each of $P=5$ prompts and calculate the mean cosine similarity of models' embeddings between each of $N \cdot P = 40$ samples and the object/style reference images. When generating an image, we do it with $50$ denoising diffusion steps. 

When investigating model performance in a continual learning regime, we use two metrics: average score \(\overline{S}_T = \frac{1}{T} \sum_{j=1}^{T} S_{j}^{T}\) (with $S_{j}^i$ being performance on $j$-th task of a model that was trained on $i$ consecutive tasks) and average forgetting \(\overline{F}_T = \frac{1}{T-1} \sum_{j=1}^{T-1} \max_{1 \leq k \leq T} (S_{j}^{k} - S_{j}^{T})\), with $S_{j}^T$ being score (CLIP-I, DINO, CSD) on $j$-th task of a model that was trained on $T$ consecutive tasks.

\section{Style personalization results}\label{appendix:results}
\begin{figure}[ht]
  \centering
  \begin{subfigure}
    \centering
    \caption{Na\"ive continual fine-tuning of LoRA adapters}
    \includegraphics[width=.6\linewidth]{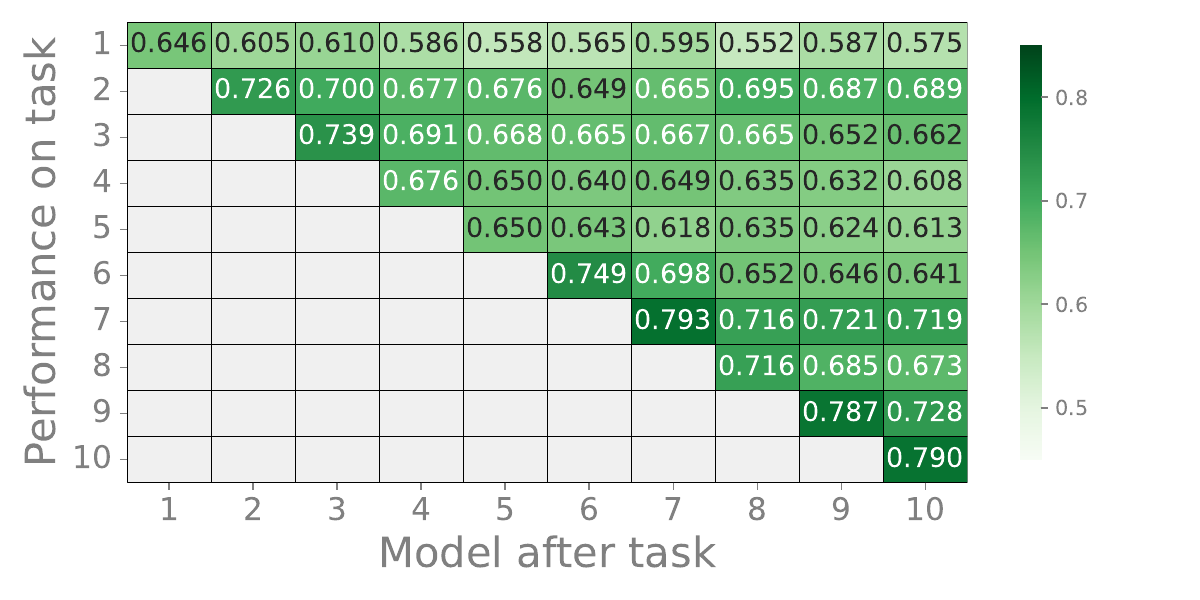}
  \end{subfigure}

  \begin{subfigure}
    \centering
    \caption{Continual adapters merging and standard LoRA reinitialization}
    \includegraphics[width=.6\linewidth]{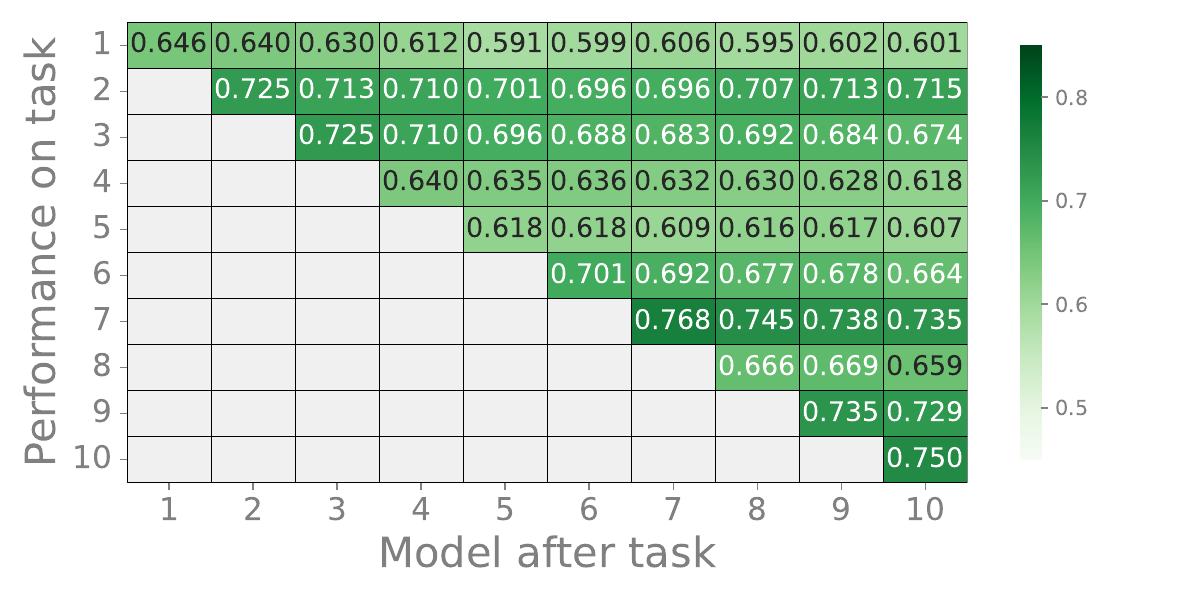}
  \end{subfigure}  
  
  \begin{subfigure}
    \centering
    \caption{Continual adapters merging and orthogonalized reinitialization}
    \includegraphics[width=.6\linewidth]{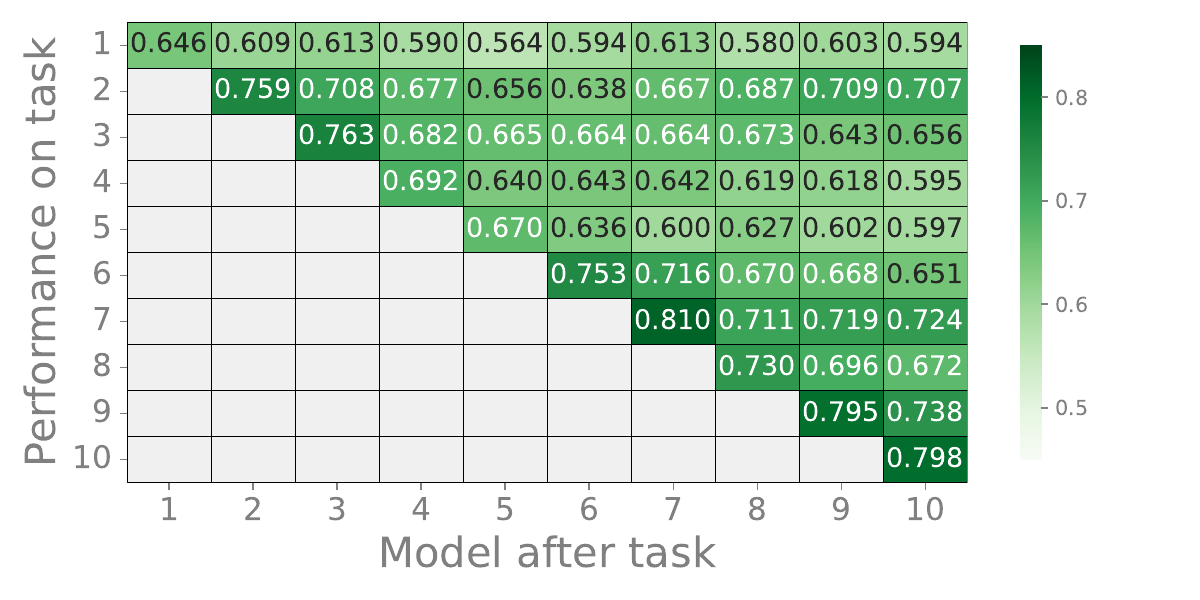}
  \end{subfigure}
  
  \begin{subfigure}
    \centering
    \caption{Magnitude-based selection of LoRA weights}
    \includegraphics[width=.6\linewidth]{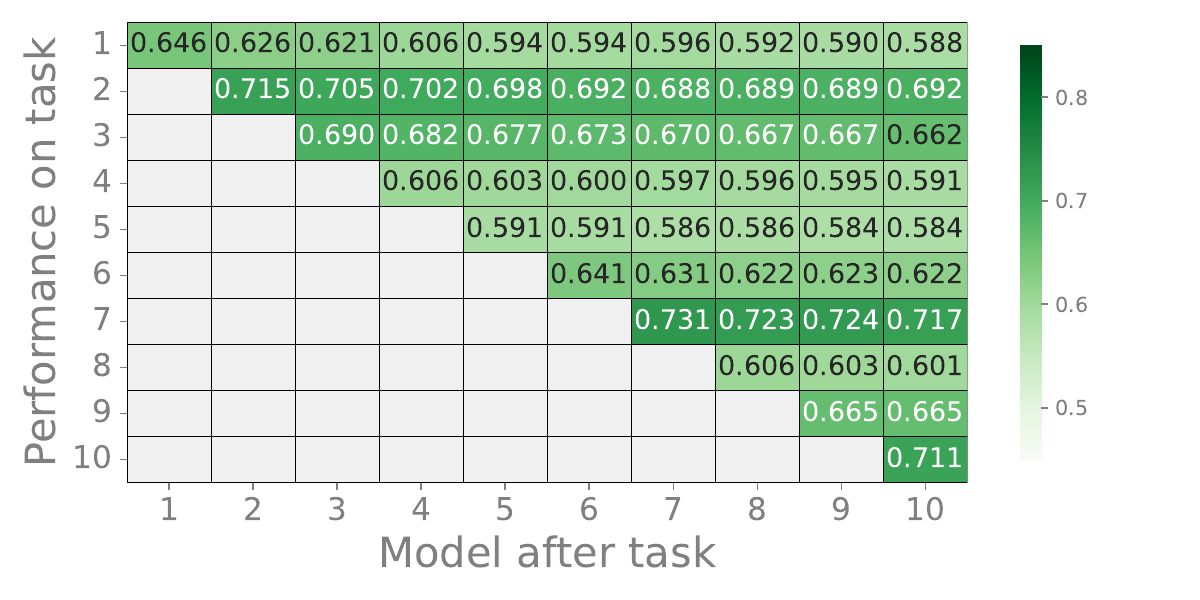}
  \end{subfigure}
  
  \caption{Heatmap of CLIP-I alignment scores for each task in continual object personalization.}
  \label{fig:heatmaps_obj_clip}
\end{figure}

\begin{figure}[ht]
  \centering
  \begin{subfigure}
    \centering
    \caption{Na\"ive continual fine-tuning of LoRA adapters}
    \includegraphics[width=.6\linewidth]{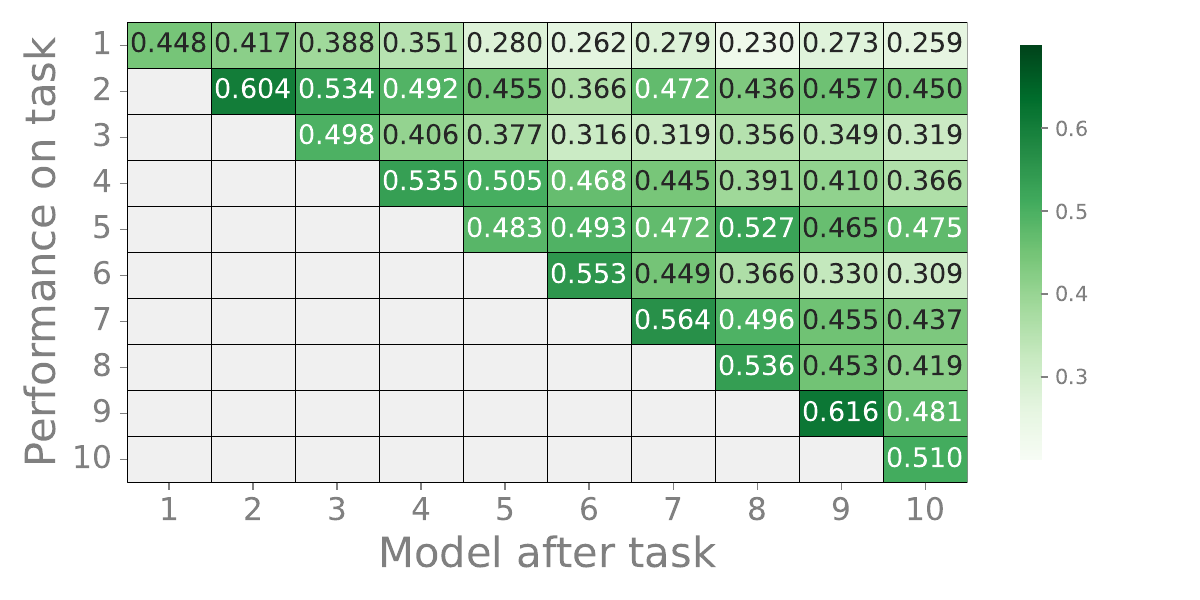}
  \end{subfigure}

  \begin{subfigure}
    \centering
    \caption{Continual adapters merging and standard LoRA reinitialization}
    \includegraphics[width=.6\linewidth]{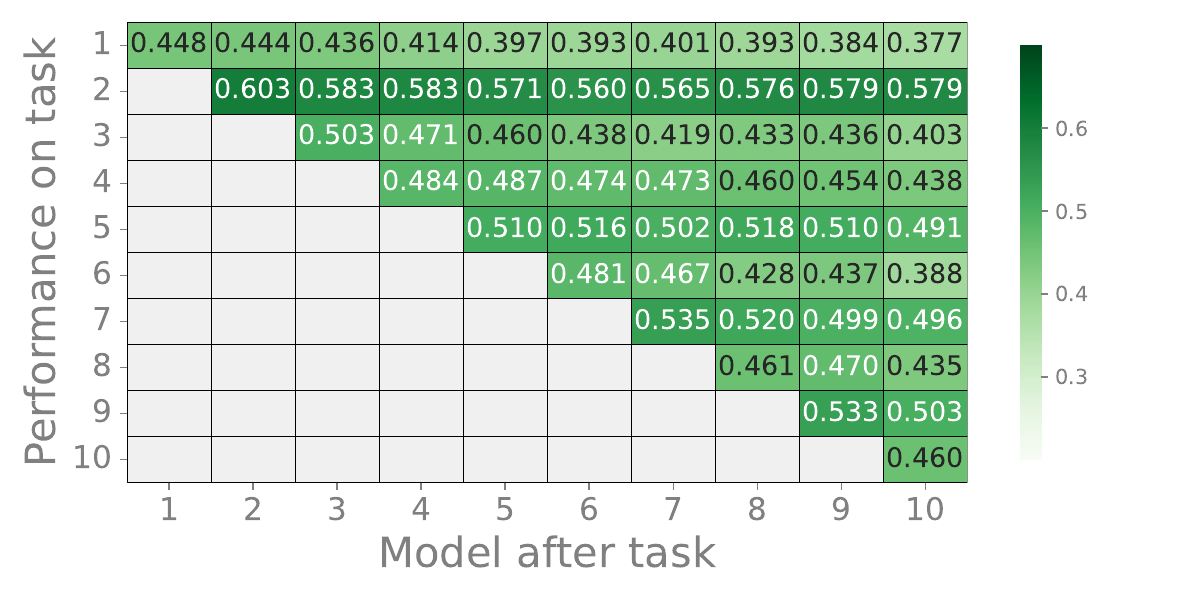}
  \end{subfigure}  
  
  \begin{subfigure}
    \centering
    \caption{Continual adapters merging and orthogonalized reinitialization}
    \includegraphics[width=.6\linewidth]{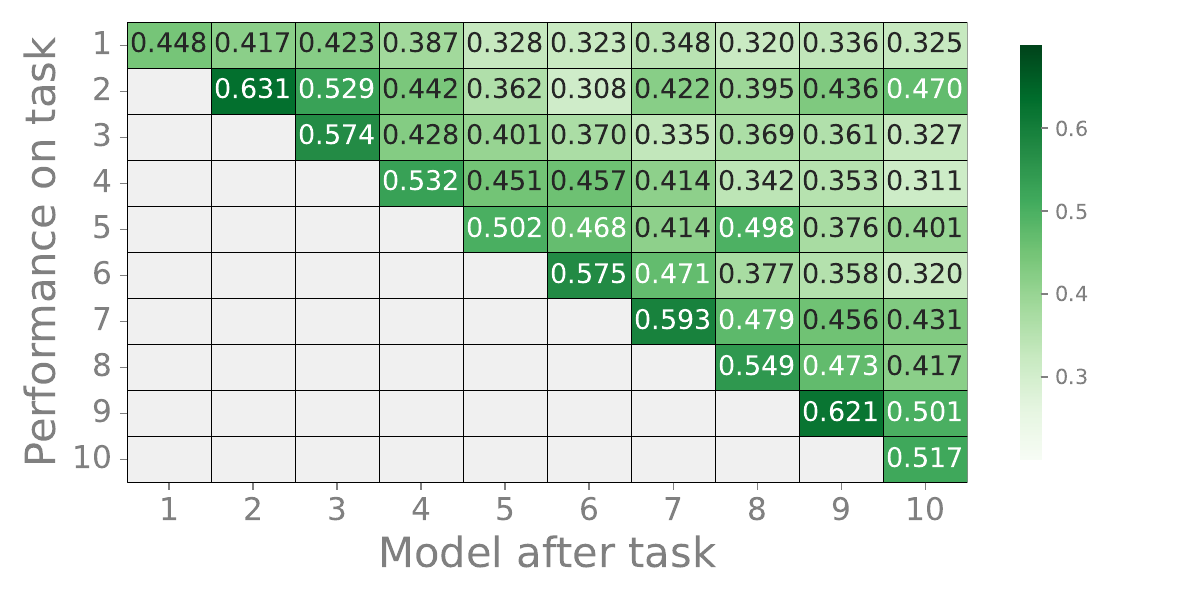}
  \end{subfigure}
  
  \begin{subfigure}
    \centering
    \caption{Magnitude-based selection of LoRA weights}
    \includegraphics[width=.6\linewidth]{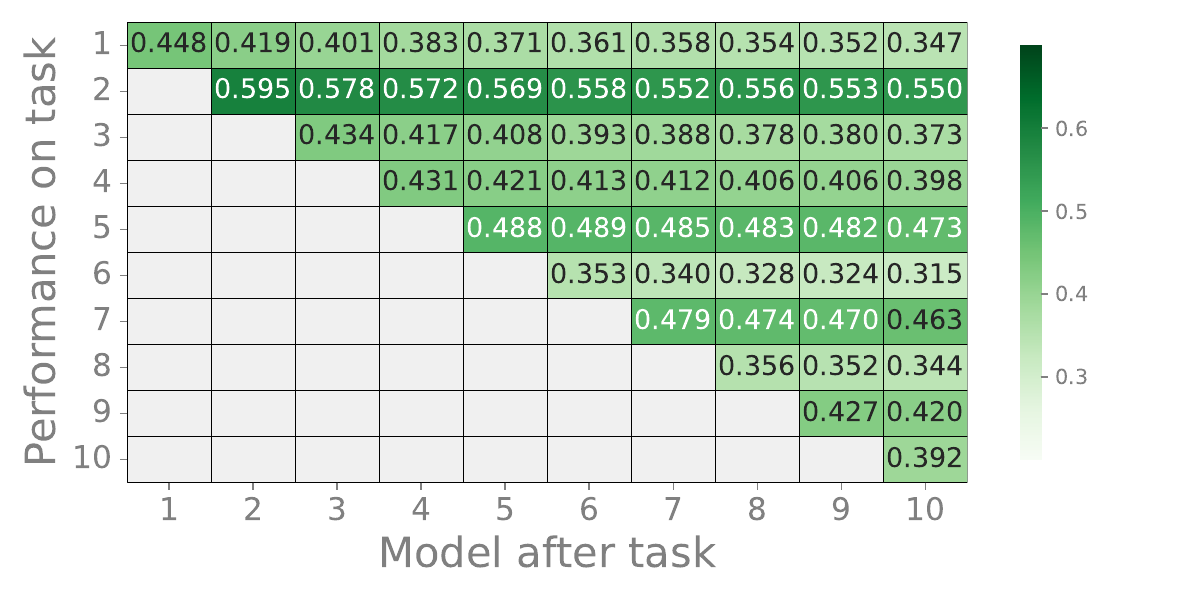}
  \end{subfigure}
  
  \caption{Heatmap of DINO alignment scores for each task in continual object personalization.}
  \label{fig:heatmaps_obj_dino}
\end{figure}

\begin{figure}[ht]
  \centering
  \begin{subfigure}
    \centering
    \caption{Na\"ive continual fine-tuning of LoRA adapters}
    \includegraphics[width=.6\linewidth]{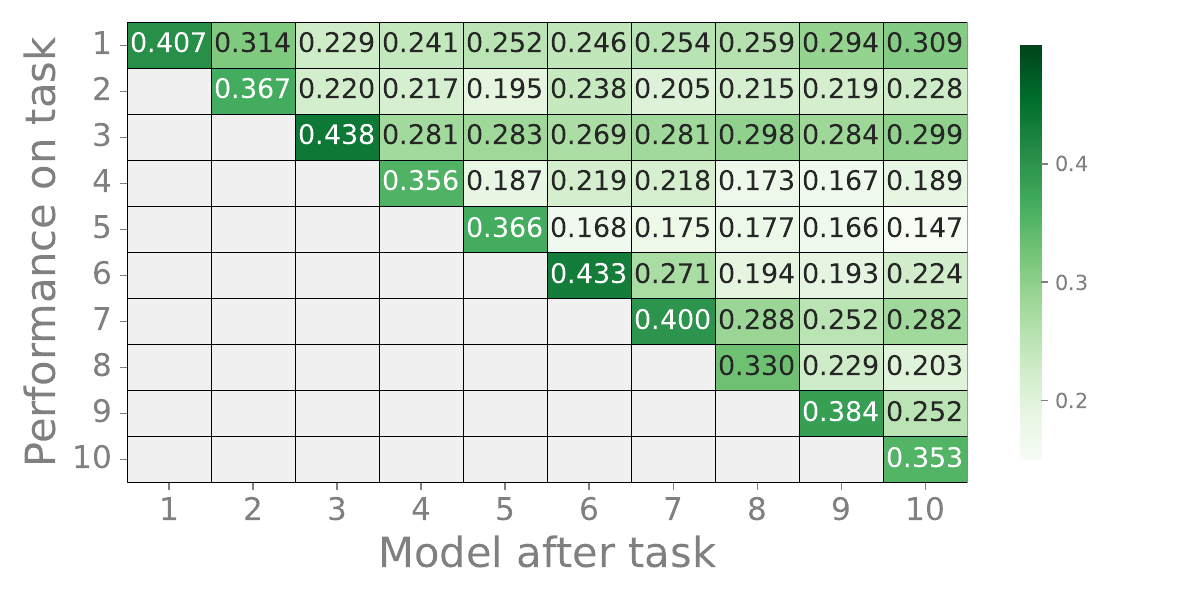}
  \end{subfigure}

  \begin{subfigure}
    \centering
    \caption{Continual adapters merging and standard LoRA reinitialization}
    \includegraphics[width=.6\linewidth]{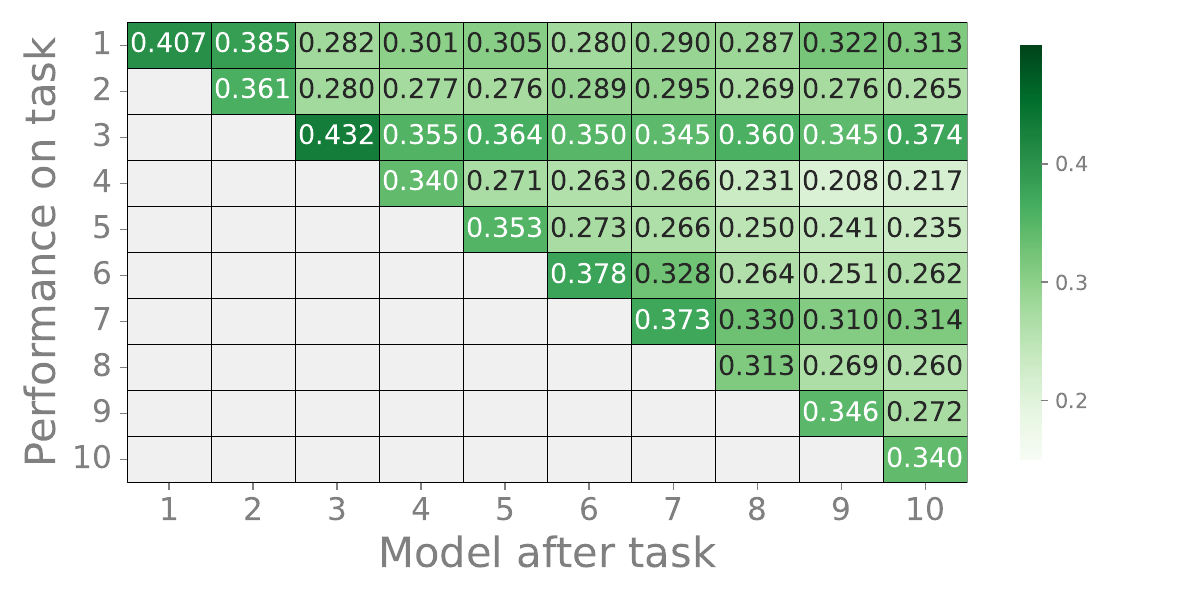}
  \end{subfigure}  
  
  \begin{subfigure}
    \centering
    \caption{Continual adapters merging and orthogonalized reinitialization}
    \includegraphics[width=.6\linewidth]{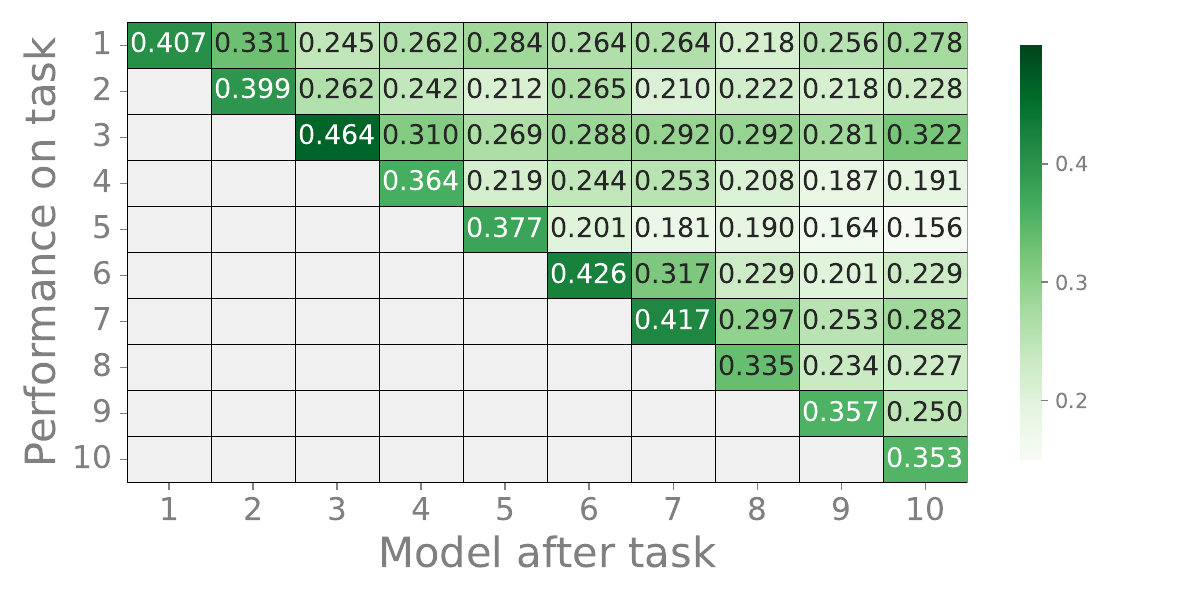}
  \end{subfigure}
  
  \begin{subfigure}
    \centering
    \caption{Magnitude-based selection of LoRA weights}
    \includegraphics[width=.6\linewidth]{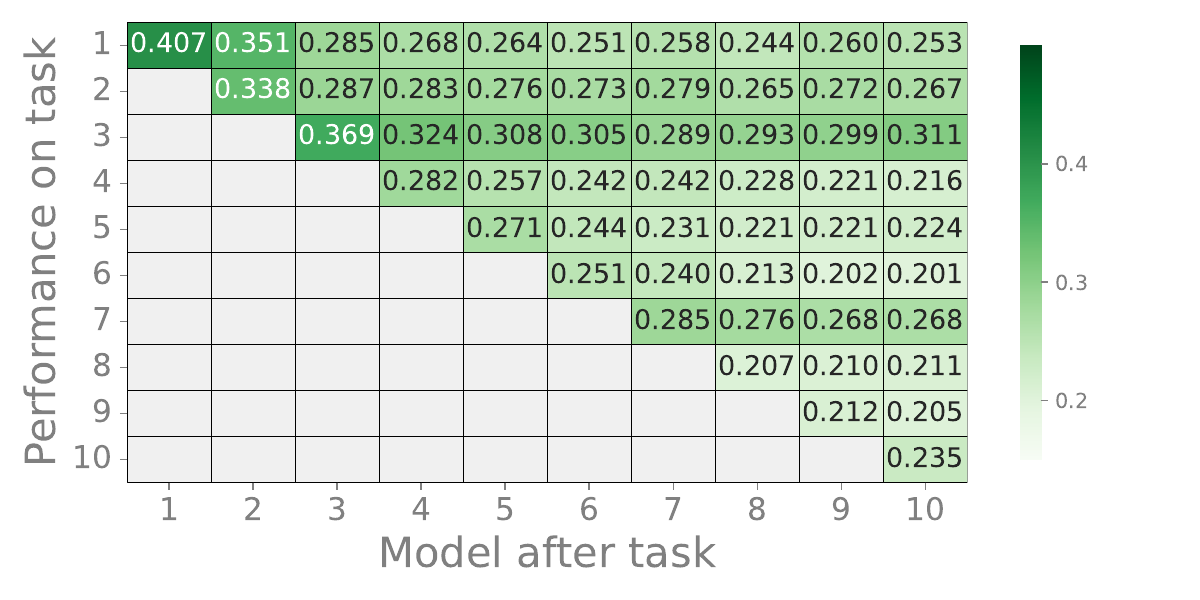}
  \end{subfigure}
  
  \caption{Heatmap of DINO alignment scores for each task in continual style personalization.}
  \label{fig:heatmaps_style_dino}
\end{figure}

\begin{figure}[ht]
  \centering
  \begin{subfigure}
    \centering
    \caption{Na\"ive continual fine-tuning of LoRA adapters}
    \includegraphics[width=.6\linewidth]{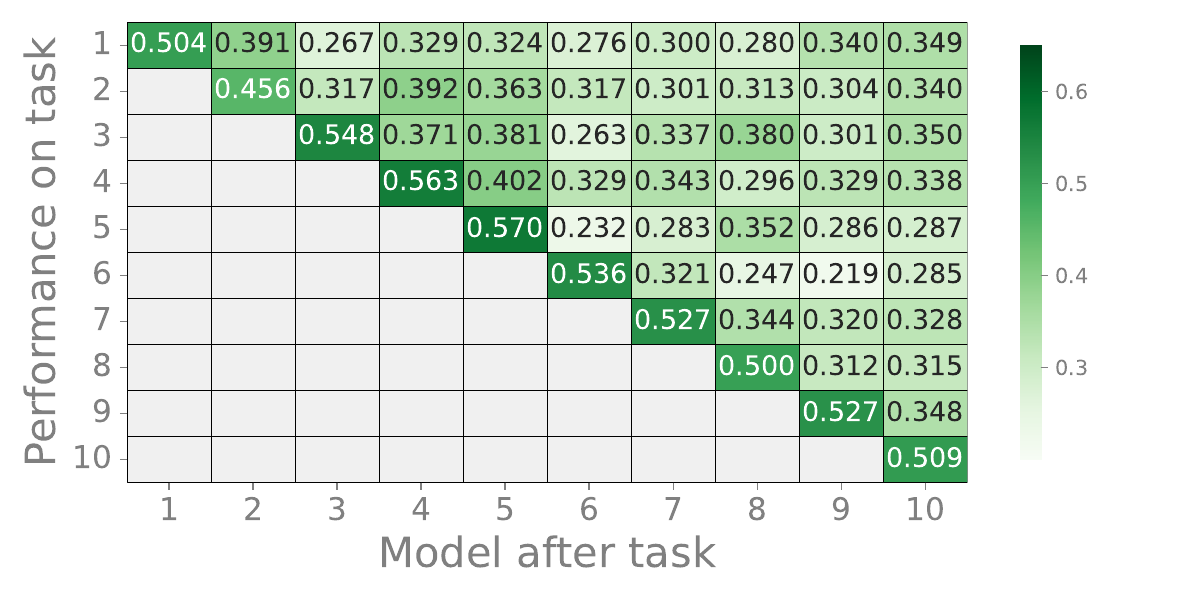}
  \end{subfigure}

  \begin{subfigure}
    \centering
    \caption{Continual adapters merging and standard LoRA reinitialization}
    \includegraphics[width=.6\linewidth]{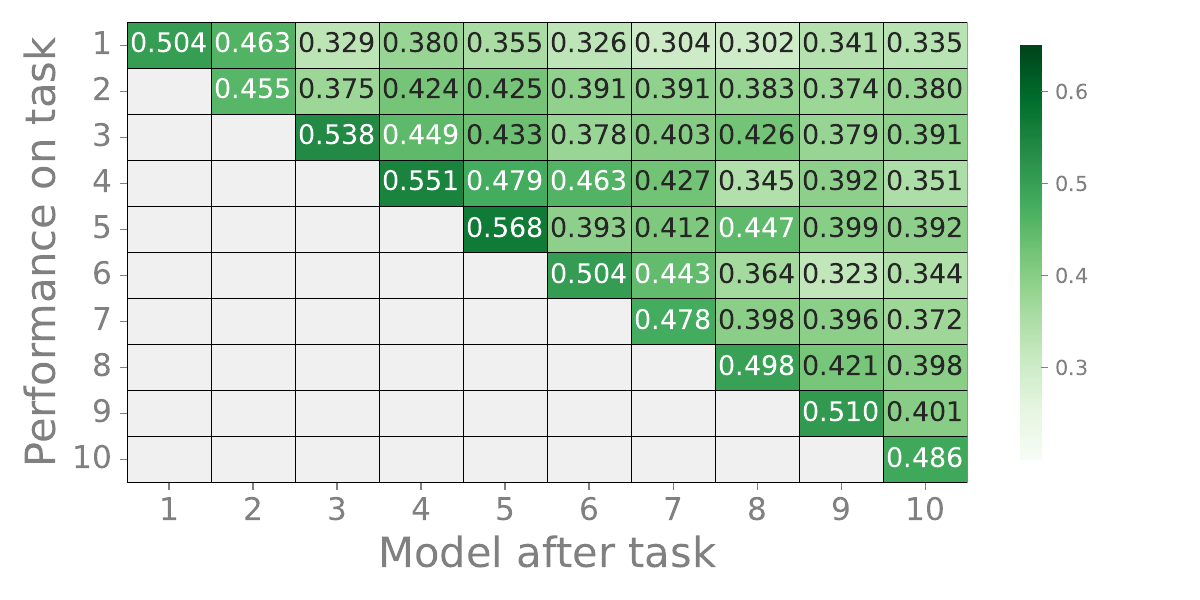}
  \end{subfigure}  
  
  \begin{subfigure}
    \centering
    \caption{Continual adapters merging and orthogonalized reinitialization}
    \includegraphics[width=.6\linewidth]{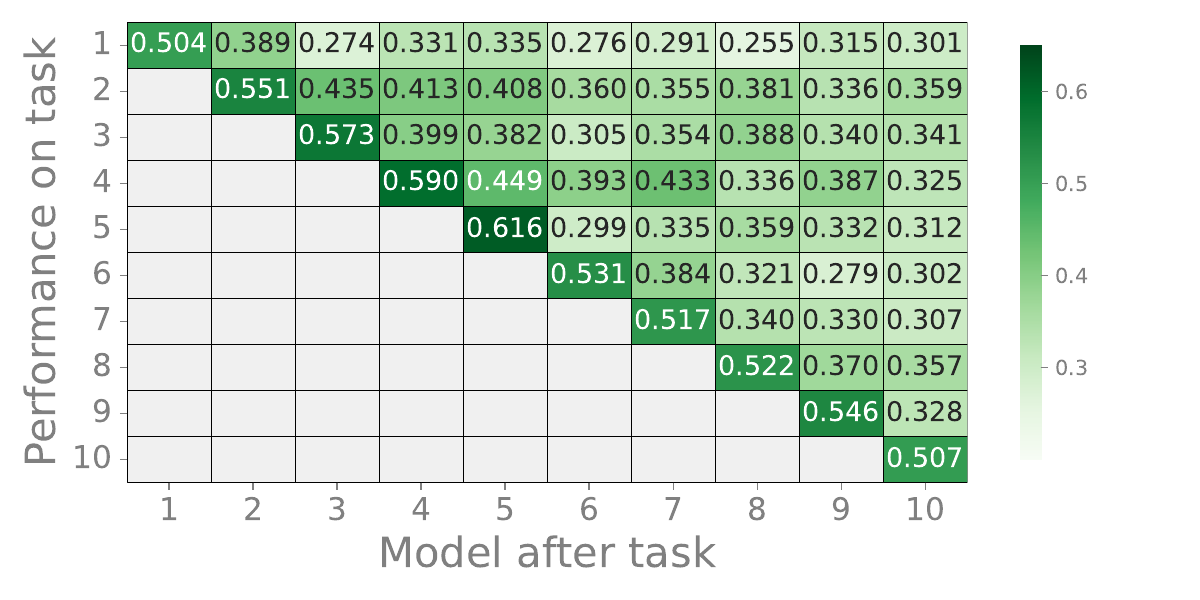}
  \end{subfigure}
  
  \begin{subfigure}
    \centering
    \caption{Magnitude-based selection of LoRA weights}
    \includegraphics[width=.6\linewidth]{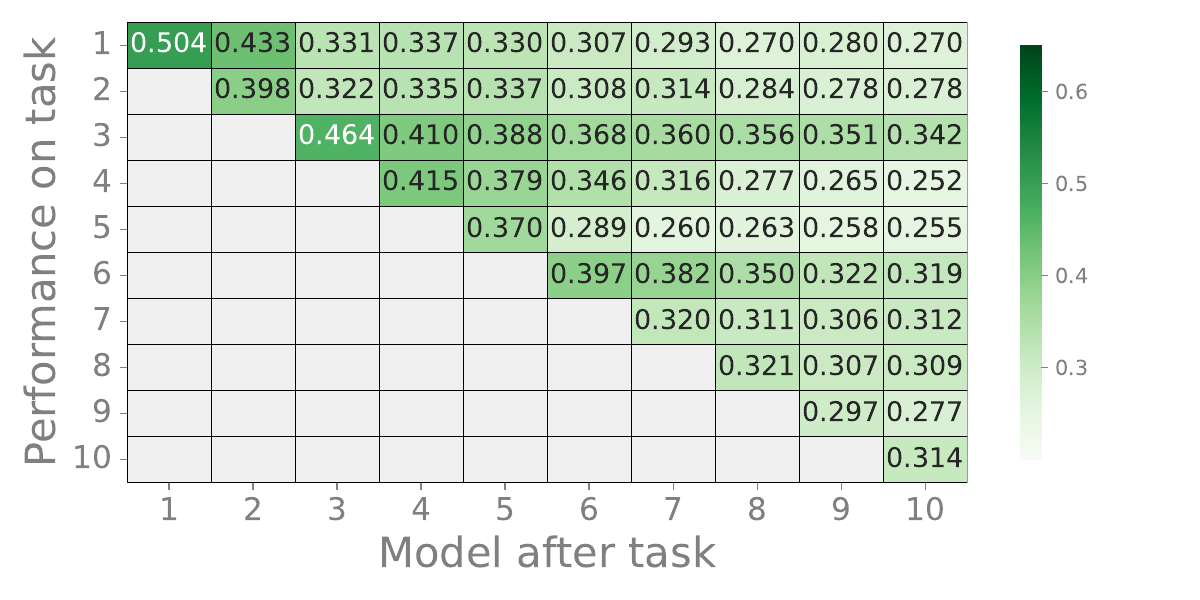}
  \end{subfigure}
  
  \caption{Heatmap of CSD alignment scores for each task in continual style personalization.}
  \label{fig:heatmaps_style_csd}
\end{figure}

The model generally achieves its highest performance on each task immediately after fine-tuning, but this performance tends to degrade as new tasks are added. Despite this decline, our evaluated techniques can retain knowledge from previous tasks. As presented in~\Cref{fig:heatmaps_obj_clip,fig:heatmaps_obj_dino} for object personalization and~\Cref{fig:heatmaps_style_dino,fig:heatmaps_style_csd} for style personalization, we can observe that the merge \& orthogonal initialization method exhibits the highest plasticity in all metrics for both objects and styles. For some cases (tasks 1 and 2 in merge \& initialization method in~\Cref{fig:heatmaps_obj_clip}), we can observe backward transfer, a situation where the performance on the previous task improves thanks to the adaptation to the new task.

\section{Prompts for styles}
Due to the complex nature of style personalization, this task is characterized by the easy overwriting of previous styles when new ones come adapted by the Dreambooth method. This fact is extremely harmful especially in the context of continued learning, where it implies catastrophic forgetting of the model. In our experiments, we note that a significant problem in the context of personalization of styles is the way they are expressed in the form of a prompt. To this end, we investigate how different forms of prompts affect the average score, average forgetting, and task scores of a diffusion model that learns the next styles sequentially. 

We evaluate models using six different prompts while training them across five consecutive style-related tasks, aiming to identify the prompt that achieved the highest accuracy and lowest forgetting. The results are summarized in~\Cref{table:style-prompts-metrics}. We also measure the average CSD alignment score and its changes across successive tasks - results in~\Cref{table:style-prompts-csd}. We observe that using classic Dreambooth prompts like 'image in the $S^*$ style' introduces worse results than specifying the full names of the styles, especially with the token 'style' inside.

\begin{table*}[!h]
\centering
\begin{tabular}{lcccc}
\toprule
\textbf{Prompt Template}               & \textbf{DINO $\overline{S}_T \, (\uparrow)$} & \textbf{DINO $\overline{F}_T \, (\downarrow)$} & \textbf{CSD $\overline{S}_T \, (\uparrow)$} & \textbf{CSD $\overline{F}_T \, (\downarrow)$} \\ 
\midrule
image of \{\} in \{token\} style   & $$0.188$$          & \underline{$$0.215$$}  & $$0.237$$          & \textbf{$$0.272$$}     \\ 
image of \{\} in \{name\} style  & $$0.289$$          & $$0.249$$              & $$0.418$$          & $$0.359$$              \\ 
image of \{\} in \{token\}   & $$0.260$$          & $$0.279$$              & $$0.377$$          & $$0.433$$              \\ 
image of \{\} in \{name\}   & \underline{$$0.292$$} & $$0.237$$            & \underline{$$0.423$$}  & $$0.348$$          \\ 
\{token\} image of \{\}   & $$0.268$$          & $$0.249$$              & $$0.387$$          & $$0.389$$              \\ 
\{name\} image of \{\}    & \textbf{$$0.299$$}  & \textbf{$$0.201$$}  & \textbf{$$0.446$$}  & \underline{$$0.275$$}  \\ 
\bottomrule
\end{tabular}
\caption{Performance metrics average score and average forgetting across different style prompts templates for DINO and CSD.}
\label{table:style-prompts-metrics}
\end{table*}

\begin{table*}[!h]
\centering
\begin{tabular}{lccccc}
\toprule
\textbf{Prompt Template}               & $\mathbf{Task_1}$ & $\mathbf{Task_2}$ & $\mathbf{Task_3}$ & $\mathbf{Task_4}$ & $\mathbf{Task_5}$ \\ 
\midrule
image of \{\} in \{token\} style   & $0.365$   &  $0.267$ &  $0.213$  &   $0.117$ & $0.084$ \\ 
image of \{\} in \{name\} style  & $\textbf{0.503}$ & $0.362$  &  $0.321$ & $0.186$ &  $0.261$ \\ 
image of \{\} in \{token\}   &   $0.491$      &   $0.303$ & $0.236$ & $0.115$  & $0.168$ \\ 
image of \{\} in \{name\}  &   $0.493$ & $ 0.371$ & $0.337$ & $0.191$ &  $0.261$ \\ 
\{token\} image of \{\}   &  $0.487$ & $0.350$ & $0.250$ & $0.130$ & $0.231$ \\ 
\{name\} image of \{\}   &   $0.487$ & $\textbf{0.410}$  & $\textbf{0.397}$ & $\textbf{0.255}$  & $\textbf{0.289}$ \\  
\bottomrule
\end{tabular}
\caption{The average value of the CSD metric demonstrates how models perform on the selected tasks, where 1 represents the most recently trained task, and 5 represents the earliest task.}
\label{table:style-prompts-csd}
\end{table*}

\section{Example samples}
\begin{figure}[ht]
    \centering
    \includegraphics[width=\linewidth]{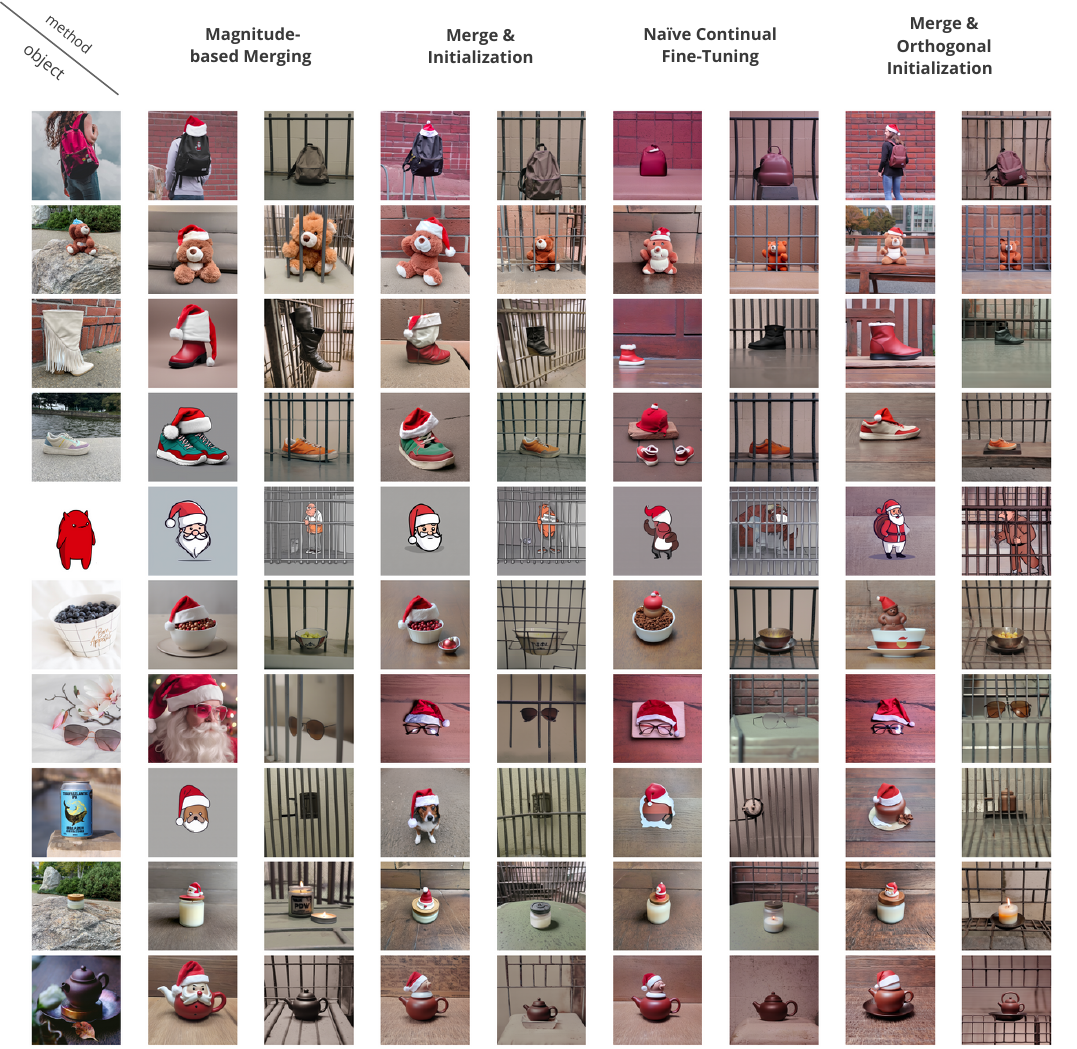}
    \caption{Images generated by the model fine-tuned on 10 tasks. The first left column represents the subsequent tasks, each represented by an image illustrating the object to learn. We show how each method sequentially performs using the prompt templates: 'a $V^*$ wearing a santa hat' and 'a $V^*$ in a jail'.}
    \label{fig:example-objects}
\end{figure}

\begin{figure}[ht]
    \centering
    \includegraphics[width=\linewidth]{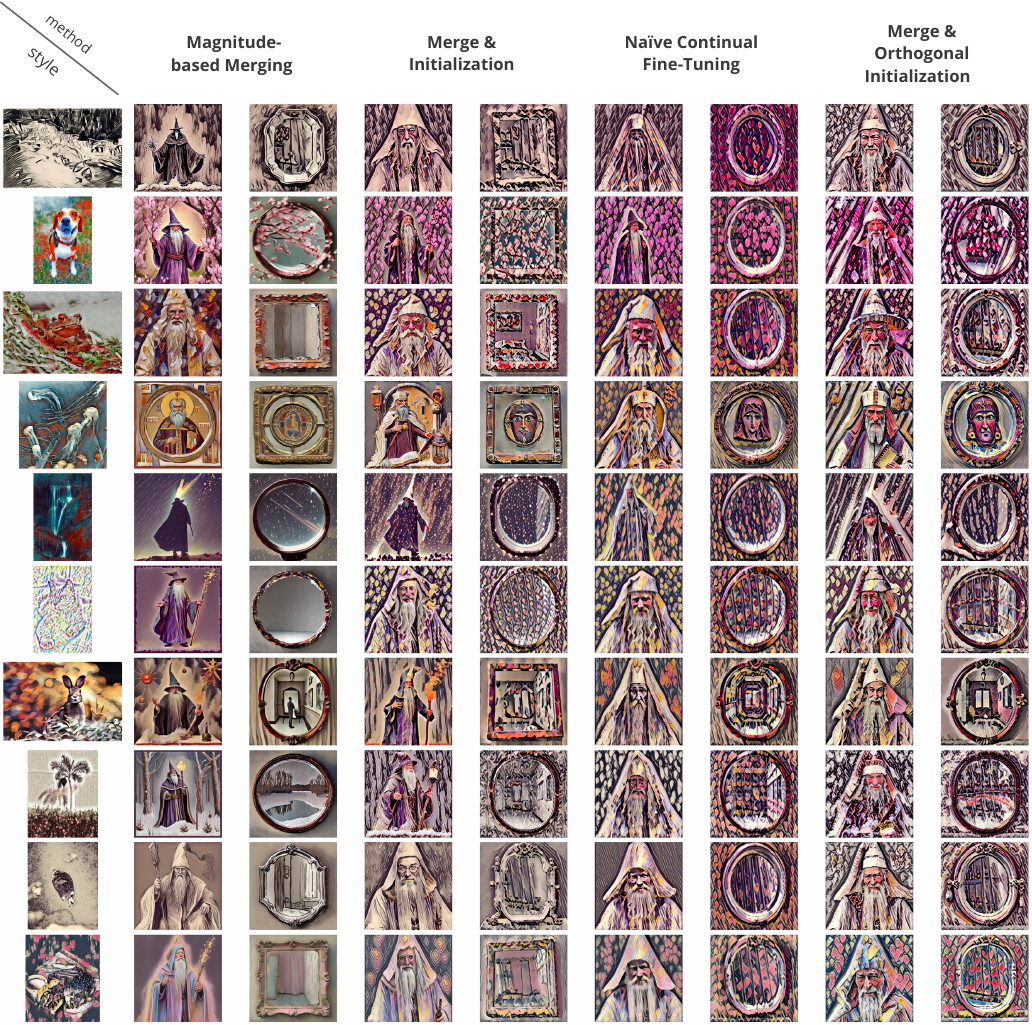}
    \caption{Images generated by the model fine-tuned on 10 tasks. The first left column represents the subsequent tasks, each represented by an image illustrating the style to learn. We show how each method sequentially performs using the prompt templates: '\{style\_name\} image of a wizard' and '\{style\_name\} image of a mirror'.}
    \label{fig:example-styles}
\end{figure}

\Cref{fig:example-objects} presents object images generated by the models fine-tuned on all 10 consecutive tasks using our continual learning methods, while~\Cref{fig:example-styles} presents it for styles.

\end{document}